%% file: main.tex
\documentclass[conference, 10pt]{IEEEtran}
\makeatletter
\def\ps@headings{%
\def\@oddhead{\mbox{}\scriptsize\rightmark \hfil \thepage}%
\def\@evenhead{\scriptsize\thepage \hfil \leftmark\mbox{}}%
\def\@oddfoot{}%
\def\@evenfoot{}}
\makeatother
\usepackage{times}
\usepackage[dvips]{graphicx}

\usepackage{euler}
\usepackage{amsmath,amssymb,amsthm}
\usepackage{epsfig,subfigure}
\usepackage[section]{algorithm}
\usepackage{algorithmic}
\usepackage{comment}
\usepackage{cite}
\usepackage{geometry}
\usepackage{color}
\usepackage{subfigure}
\usepackage{epstopdf}

\usepackage{geometry}
\usepackage{epstopdf}

\begin{document}

\title{Automatic Objects Removal for Scene Completion}

%\author{Jianjun Yang $\;\; $Ju Shen$\;\;$  Yin Wang $\;\;$ Honggang Wang  $\;\;$ Kun Hua $\;\;$ Wei Wang
%
%}

\author{
\IEEEauthorblockN{Jianjun Yang}
\IEEEauthorblockA{Department of Computer Science\\
University of North Georgia\\
Oakwood, GA 30566, USA\\
Email: jianjun.yang@ung.edu}\\[0.5cm]  %<------ Line breaks in the current column

\IEEEauthorblockN{Kun Hua}
\IEEEauthorblockA{Department of Electrical and \\
Computer Engineering\\
Lawrence Technological University\\
Southfield, Michigan 48075, USA\\
Email: khua@ltu.edu}

\and
\IEEEauthorblockN{Yin Wang}
\IEEEauthorblockA{Department of Mathematics and \\
Computer Science\\
Lawrence Technological University\\
Southfield, Michigan 48075, USA\\
Email: ywang12@ltu.edu}\\[0.3cm]  %<------- Extra vertical space

\IEEEauthorblockN{Wei Wang}
\IEEEauthorblockA{Department of Electrical Engineering\\ and Computer Science\\
South Dakota State University, USA\\
Email: wei.wang@ieee.org}

\and

\IEEEauthorblockN{Honggang Wang}
\IEEEauthorblockA{Department of Electrical \\
and Computer Engineering\\ %%Engineering\\
University Massachusetts Dartmouth, USA\\
Email: hwang1@umassd.edu}\\[0.5cm]           %<-----------

\IEEEauthorblockN{Ju Shen}
\IEEEauthorblockA{Department of Computer Science\\
University of Kentucky\\
Lexington, KY 40506, USA\\
Email: jushen.tom@uky.edu}
}

\maketitle
\thispagestyle{empty}

%\begin{abstract}
%\input{abs}
%\end{abstract}
%\vspace{-0.05in}

%\begin{keywords}
%  Probability, scheduling, data mining, prefetch, heterogeneous
%\end{keywords}

\begin{abstract}
\input{abs}

\end{abstract}

%\noindent
%{\bf\normalsize ABSTRACT}\newline
%{
%\input{abs}
%}
%\vspace{2ex}
%
%
%\noindent
%{\bf\normalsize KEY WORDS}\newline
%{Probability, scheduling, data mining, prefetch, heterogeneous}

\vspace{2ex} {\bf\normalsize Keywords:} {Image Completion, Texture Synthesis, On-line Photos, Scene Reconstruction, Object Removal}

\input{intro}

\input{rel}

\input{alg}

\input{exp}
\input{q_con}

\bibliographystyle{IEEE}

\bibliography{ref}

%\bibliographystyle{IEEEbib}
%\bibliography{ref}

\end{document}

%% file: abs.tex
%Low energy consumption
With the explosive growth of web-based cameras and mobile devices, billions of photographs are uploaded to the internet. We can trivially collect a huge number of photo streams for various goals, such as 3D scene reconstruction and other big data applications.  However, this is not an easy task due to the fact the retrieved photos are neither aligned nor calibrated. Furthermore, with the occlusion of unexpected foreground objects like people, vehicles, it is even more challenging to find feature correspondences and reconstruct realistic scenes. In this paper, we propose a structure-based image completion algorithm for object removal that produces visually plausible content with consistent structure and scene texture. We use an edge matching technique to infer the potential structure of the unknown region. Driven by the estimated structure, texture synthesis is performed automatically along the estimated curves. We evaluate the proposed method on different types of images: from highly structured indoor environment to the natural scenes. Our experimental results demonstrate satisfactory performance that can be potentially used for subsequent big data processing: 3D scene reconstruction and location recognition.

%Image in-painting is a popular and powerful technique for fixing undesirable content on digital photographs. The challenges of this technique are to estimate the unknown content while preserving the spacial and visual consistency.
%In this paper, we proposed a novel algorithm for image completion that could automatically produce visually plausible content with consistency in both structure and texture. The idea of this approach is to infer the potential structure in the missing region before texture generation. Our experiments show that this approach achieve favorable result than existing state-of-art work.
%
%Using features (e.g. SIFT) to automatically estimate correspondence information and reconstruct 3D geometry from the internet photos.
%And even for hard real-time, we have good results.

%% file: intro.tex
\section{Introduction}
\label{sec:intro}
%\vspace{-0.1in}

\begin{figure*}
    \centering
    \subfigure[]
    {
        \includegraphics[width=5.0cm]{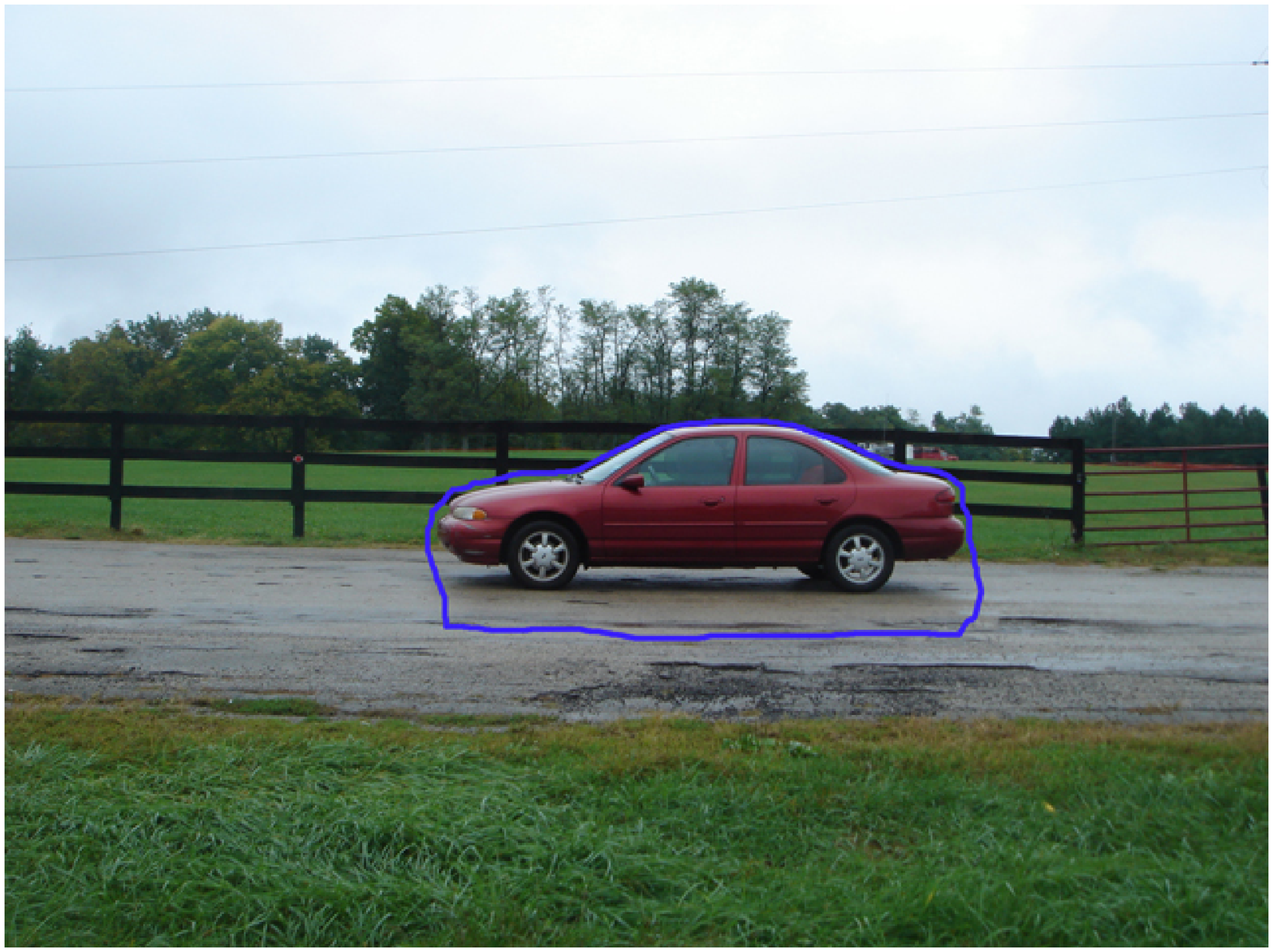}
        \label{fig:demo a}
    }
    \subfigure[]
    {
        \includegraphics[width=5.0cm]{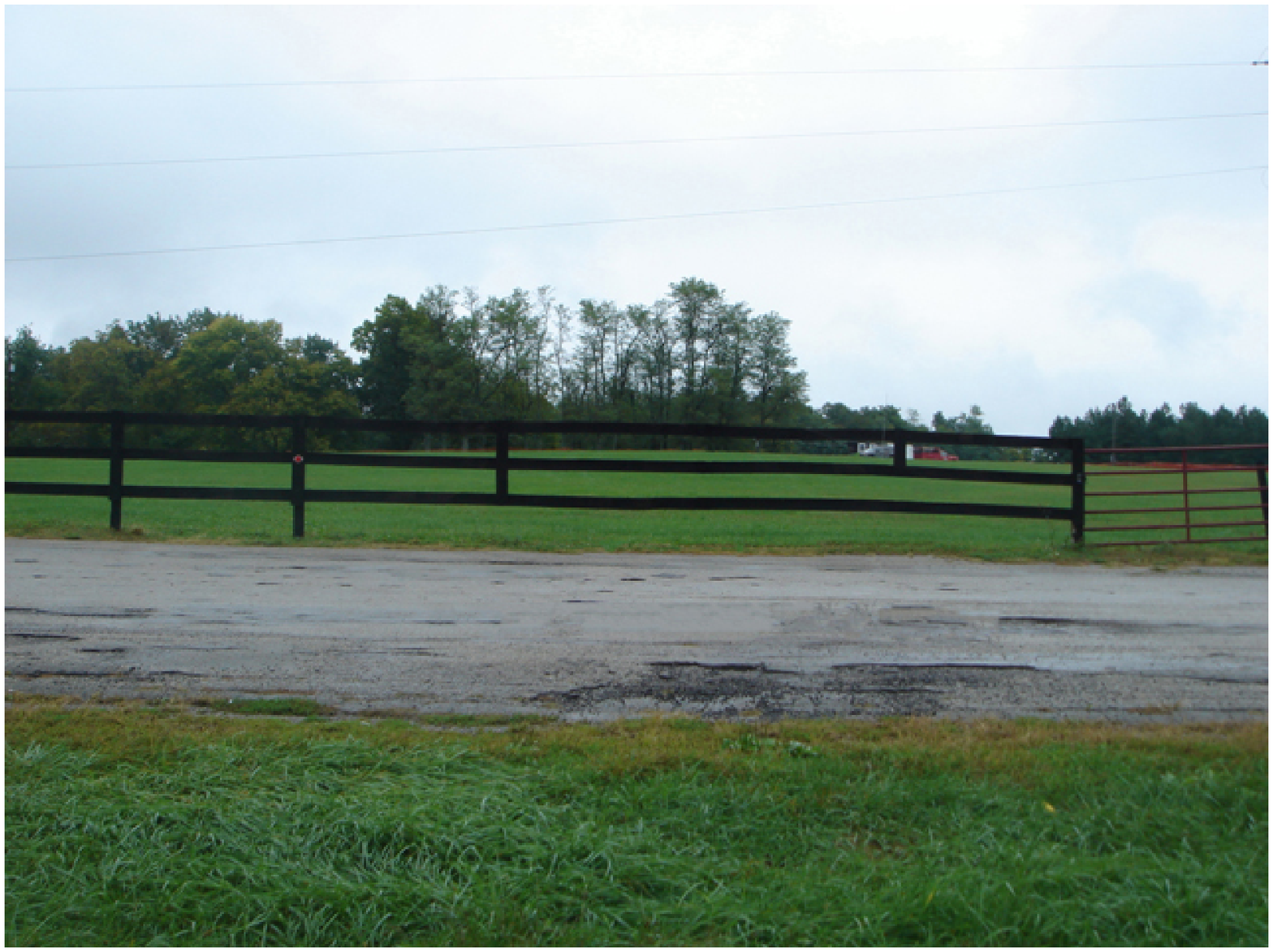}
        \label{fig:demo b}
    }
     \subfigure[]
    {
        \includegraphics[width=5.0cm]{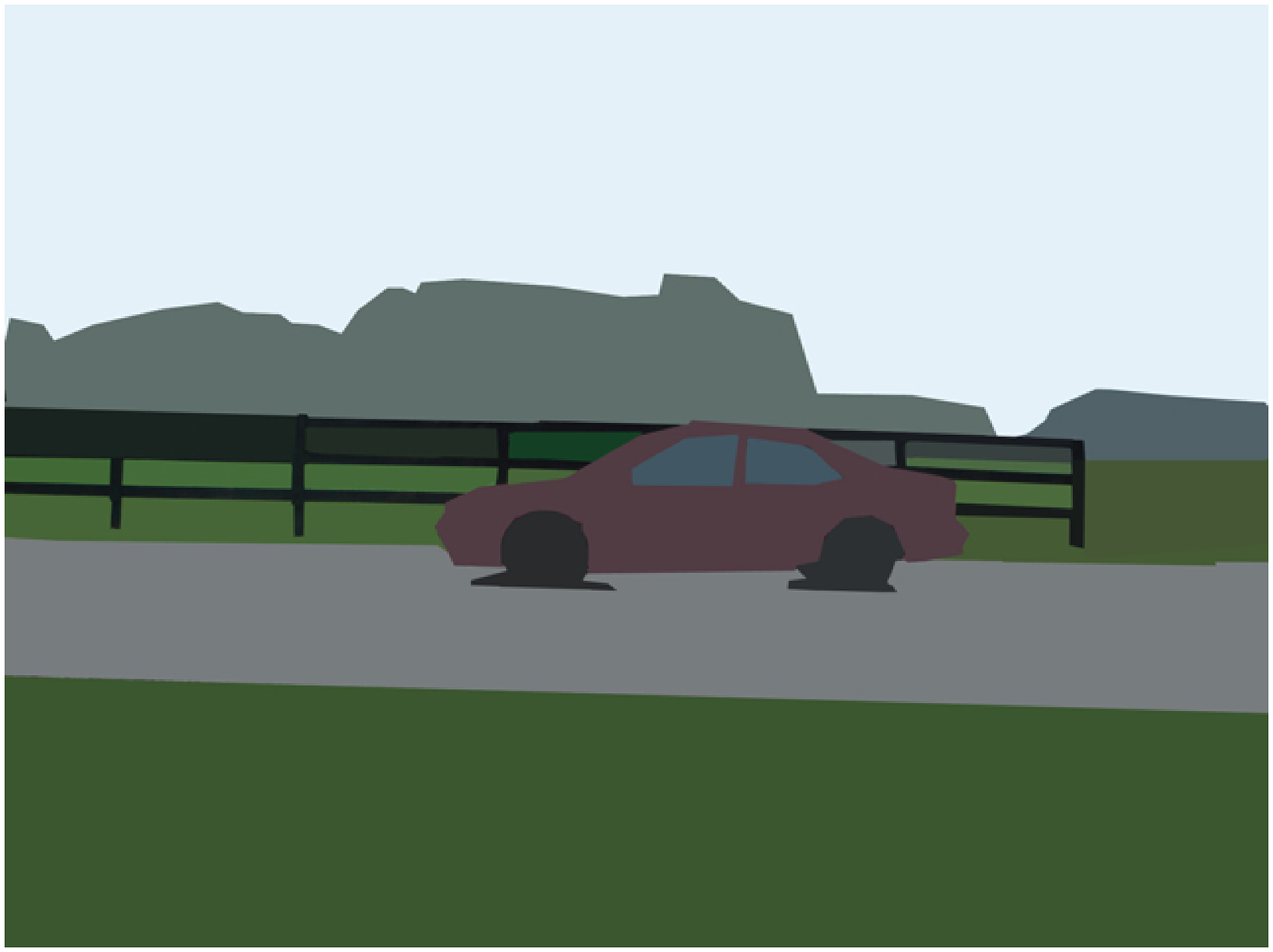}
        \label{fig:contour a}
    }
    \subfigure[]
    {
        \includegraphics[width=5.0cm]{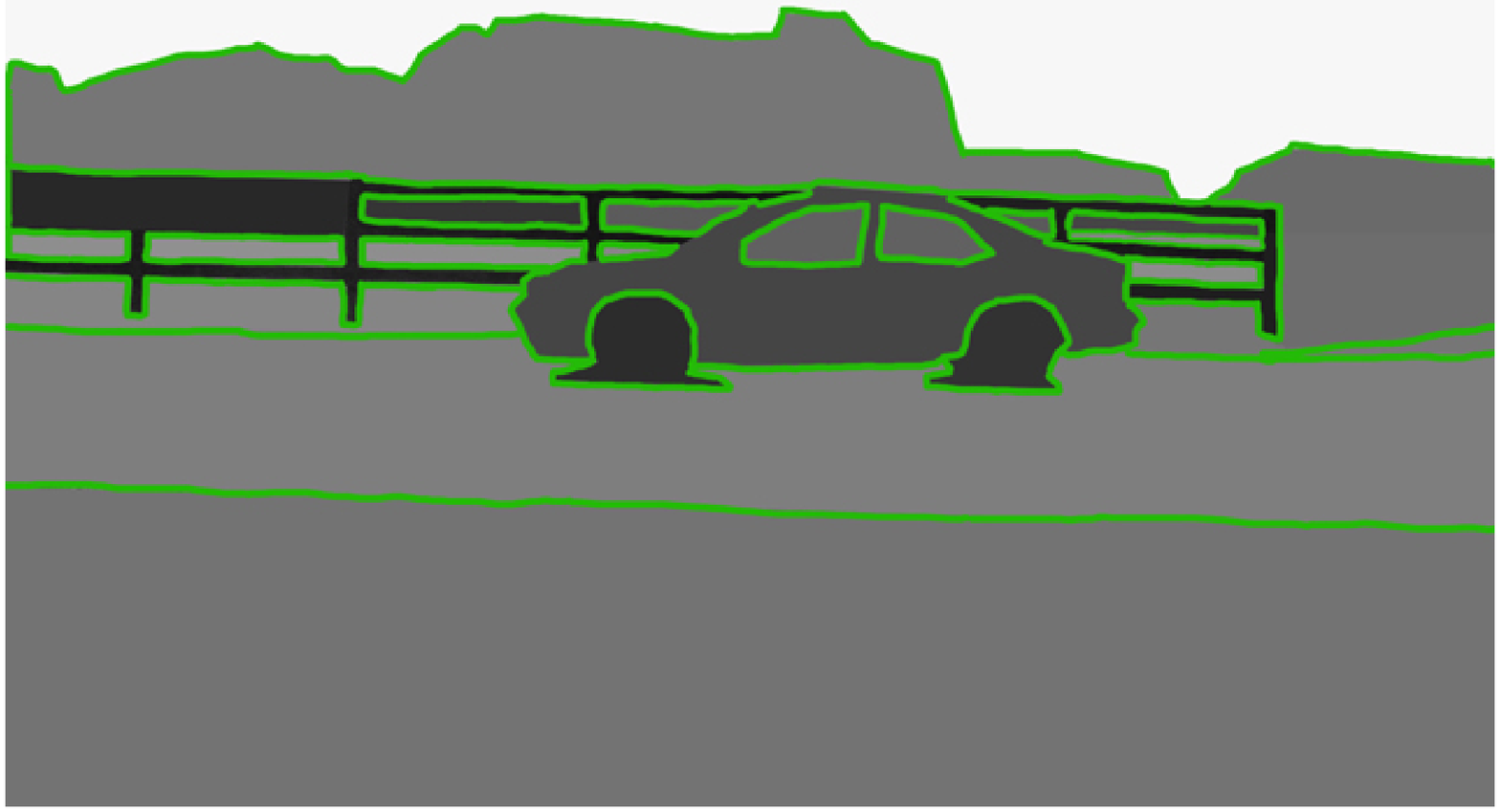}
        \label{fig:contour b}
    }
    \subfigure[]
    {
        \includegraphics[width=5.0cm, height=3.7cm]{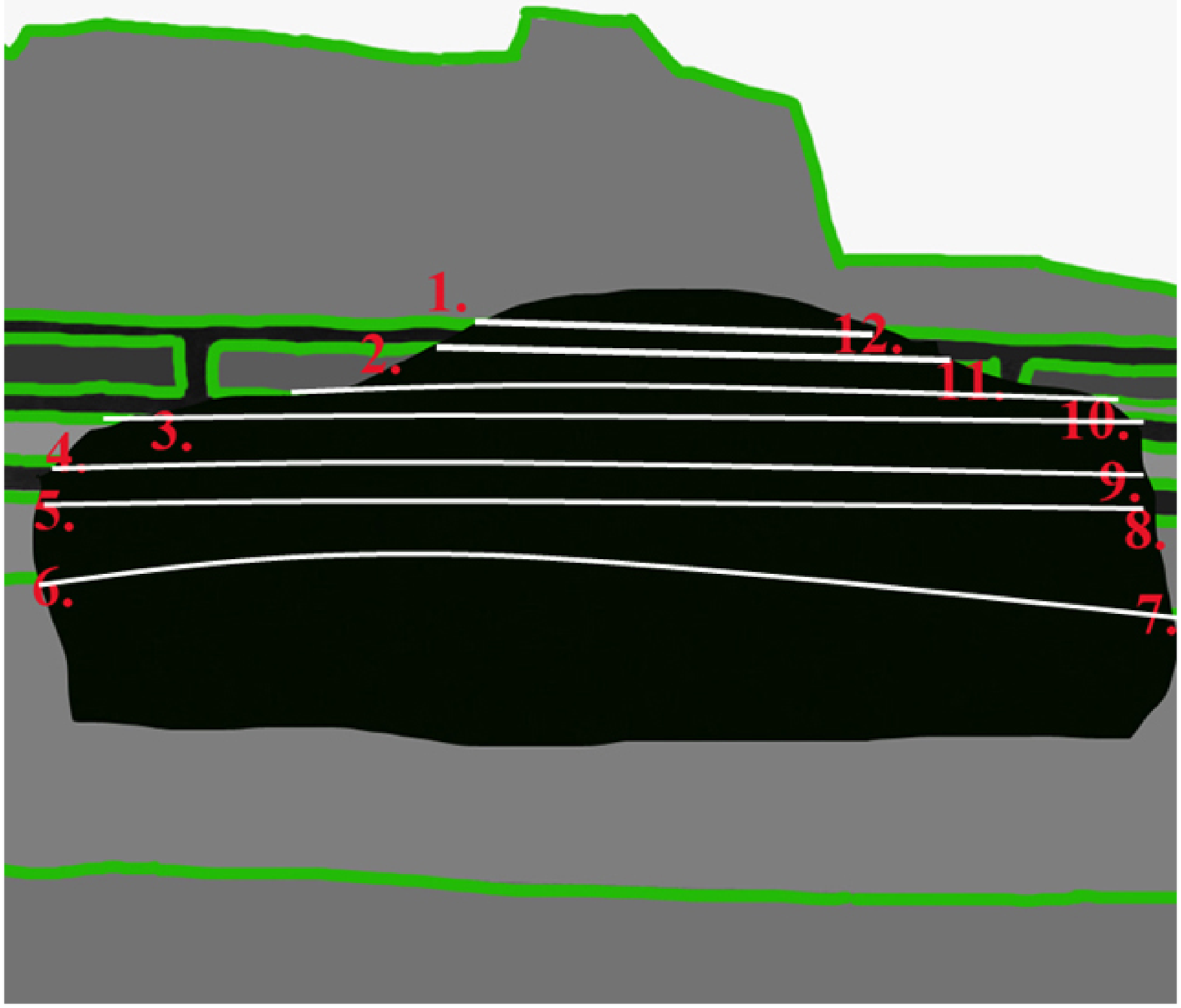}
        \label{fig:angle}
    }
    \subfigure[]
    {
        \includegraphics[width=5.0cm, height=3.7cm]{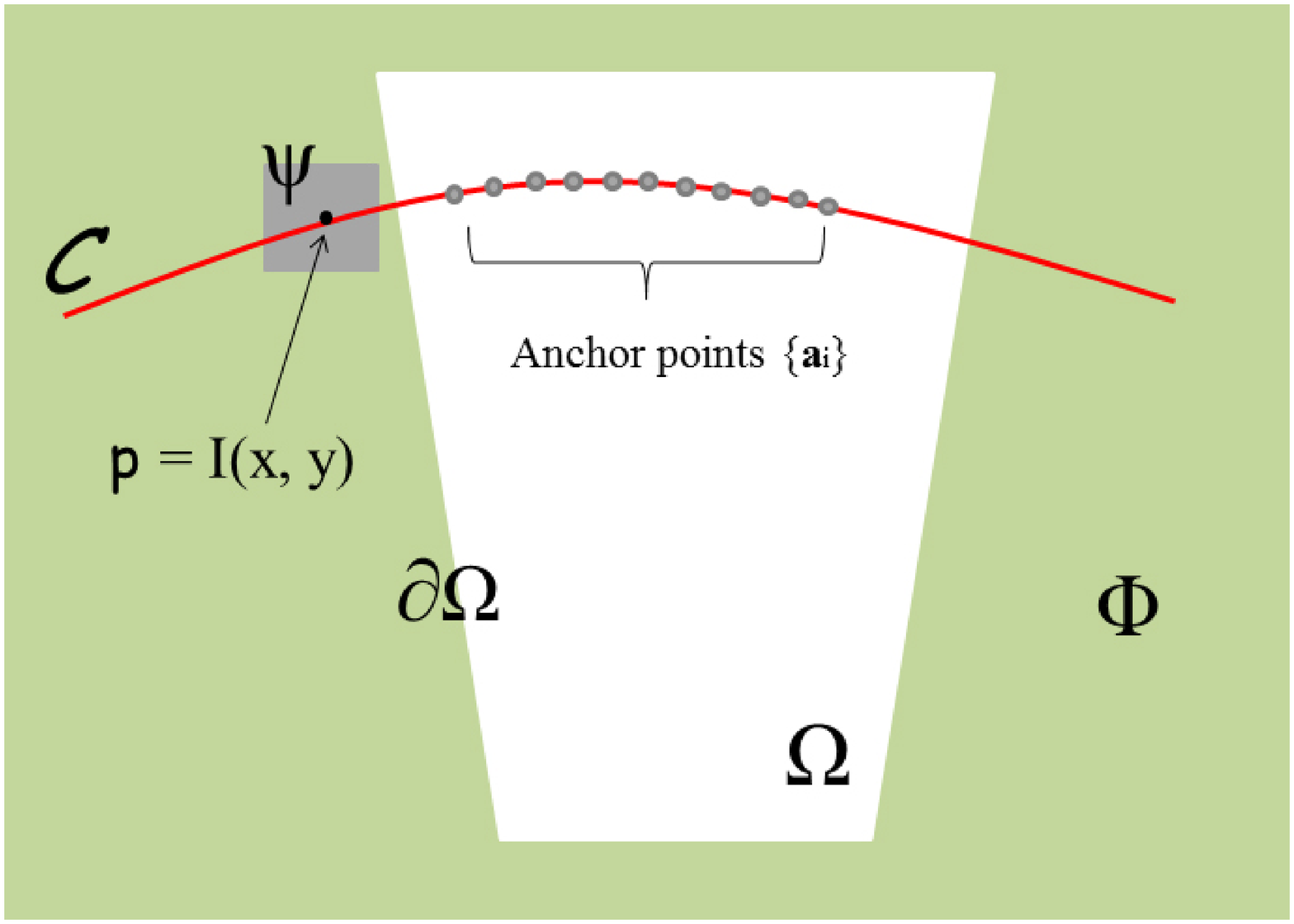}
        \label{fig:symbols}
    }
    \caption{Scene recovery by removing specified foreground object  (a) Original Image (b) Our result (c) Contour detection by using \emph{OWT-UCM} method \cite{Arbel_06} (d) Edge extraction (e) Structure generation in the occlusion region by identifying corresponding edge pairs (f) Some denotations in our algorithm. }
    \label{fig:demo}
\end{figure*}

In the past few years, the massive collections of imagery on the Internet have inspired a wave of work on many interesting big data topics: scene reconstruction, location recognition, and online sharing of personal photo streams\cite{su2013sdquery}\cite{wang2013supporting}\cite{wang2012scimate}. For example, one can easily download a huge
number of photo streams associated with a particular place. By using features (e.g. SIFT), it is possible to automatically estimate correspondence information and reconstruct 3D geometry for the scene\cite{sun2012unsupervised}\cite{hu2010trustworthy}. Imagine building a world-scale location recognition engine from all of the geotagged
images from online photo collections, such as Flickr and street view databases from Google and Microsoft. However, it is a challenging task as the photo streams are neither aligned nor calibrated since they are taken in different temporal, spatial, and personal perspectives. Furthermore, with the occlusion of unexpected foreground objects, it is even more difficult to recover the whole scene or accurately identify overlapping regions between different photos. 
%Taking the Google street view as an example, the passing vehicles or walking passengers introduce many noticeable artifacts during the image %transformation. These moving objects also potentially affect the accuracy in image localization and view reconstruction.

To resolve the above issue, image in-painting is an effective solution. 
%that has been widely used in many fields from deteriorated digital photograph reconstruction to objects removal in graphic design. 
In this paper, we propose an automatic object removal algorithm for scene completion, which benefits subsequent large imagery processing. The core of our method is based on the structure and texture consistency.
%First, it predicts the underlying structure of the occluded region by edge detection and contour analysis. Then structure propagation is applied to the region followed by a patch-based texture synthesis. 
Our proposed approach has two major contributions. First, we develop a curve estimation approach to infer the potential structure of the occluded region on the image.
Second, an orientated patch matching algorithm is designed for texture propagation. 
%Our experiments demonstrate satisfactory results. Starting with such structure based scene completion by removing foreground objects, rather than a collection of raw images, 
Our work has a broad range of applications including image localization\cite{yang2010}\cite{yang2013}, privacy protection\cite{juextra_3}\cite{juextra_4}\cite{pan2012using}, and other network based applications \cite{yang2008bipartite}\cite{yang2011statistical}\cite{xu2013efficient}\cite{xu2013energy}\cite{xu3}.

%First, by removing foreground objects in the image, the matching accuracy can be dramatically improved as the corresponding features (e.g. SIFT) are only extracted from the static scene regions rather than temporary moving objects. Second, without the existence of foreground objects in the photo, the generated views are more realistic because the foreground pixels are involved in the image transformation and geometric estimation. This work can benefit

%And even for hard real-time, we have good results.

%% file: rel.tex
\section{Related Works}
\label{sec:rel}
In the literature, image completion or in-painting has been intensively studied: in \cite{Efros_99}, Efros and Leung used a one-pass greedy algorithm to render unknown pixels based on the assumption that the probability distribution of the pixel's brightness is independent to the rest of the image when the spatial neighborhood is given. In \cite{Efros_01}, the authors proposed an example-based approach to fill in the missing regions. It worked well in filling in small gaps but not in large ones. The weakness of such approach is that it fails to preserve the potential structures.  %Bertalmio \emph{et al}. \cite{Bertalmio_03} used \emph{Partial Differential Equations} (PDE) to propagate image laplacians. The basic idea is to divide the whole image into two parts. For one part, it is completed by texture synthesis; for the other one, it is in-painted by bilinear interpolation. The two parts are then merged together as the final image. This approach often suffers from blurring pixels. As an improvement, Drori \emph{et al}. \cite{Drori_03} proposed an approach to enhance those blurring pixels while smoothening the surrounding regions. But the algorithm was too slow. 
Jia \emph{et al}. \cite{Jia_03} designed an image in-painting method based on texture-segmentation and tensor-voting that created smooth linking structures in the occluded regions. This method sometimes introduces noticeable artifact due to the texture inconsistency.  Criminisi \emph{et al}. \cite{Criminisi_04} made an improvement by assigning in-painting orders based on the edge strength levels.  Their algorithm used a confidence map and the image edges to determine the patch completion priority. However, the structures in the resulting images are not well preserved. The method in \cite{sun_05} produced a better result via structure propagation, while this approach requires more interaction. The completion results largely depend on the animator's individual technique. Some other existing work  also explored in\cite{juextra_1}\cite{juextra_2}\cite{6607541}.

%Our algorithm is derived from the following observations:spacial coherence among neighboring pixels imposes strong constraints on rendering the unknown region. The definition have been formalized mathematically in  \cite{Criminisi_04}\cite{Efros_01}\cite{Freeman_00}\cite{Efros_99}\cite{Bonet_97}\cite{c_01}\cite{Wey_00}. Exemplar-based approach is well-proved approach for this purpose. However, a naive copy-and-paste of image patches may lead to noticeable artifacts. Thus structure preservation is a key factor in solving this problem. There exists several techniques \cite{sun_05} for structures maintenance. For different images, the above two factors may play different roles in rendering the missing region. According to many previous works \cite{Criminisi_04}, the order of rendering pixels is critical the quality of resulting image.

%Image in-painting have aroused the interests of image processing researchers in the last decade. It has been widely used in many fields from deteriorated digital photograph reconstruction to objects removal in graphic design. Several algorithms and approaches have been developed to automatically remove an object in an image and fill the obscured regions or restore scratched regions. Two main categories of algorithms were used to solve the problem: texture synthesis for large gaps and in-painting for small gap.

%% file: alg.tex
\section{Our Approach}
\label{sec:alg}
%\vspace{-0.1in}
The process of our framework is: for a given image, users specify the object for removal by drawing a closed contour around it. The enclosure is considered an unknown region that is inferred and replaced by the remaining region of the image. Figure \ref{fig:demo a} shows an example: the red car is selected as the removing object. In the resulting image Figure \ref{fig:demo b}, the occluded region is automatically recovered based on the surrounding environment.

%\subsection{Definitions and Lemma}
First let us define a set of notations for the rest of our paper. For an image $\mathbf{I}$, the target region for in-painting is denoted as $\mathbf{\Omega}$; the remaining part of the image is denoted as $\mathbf{\phi} (= \mathbf{I} - \mathbf{\Omega})$, which is also known as source region. The boundary contour along $\mathbf{\Omega}$ is denoted as $\mathbf{\partial\Omega}$. A pixel's value is represented by $p = \mathbf{I(x, y)}$, where x and y are the $2D$ coordinates on the image. The surrounding neighborhood centered at $(x, y)$ is often called as a patch, denoted as $\Psi_p$. The coordinates of pixels inside the patch $\Psi_p$ should be in the range: $[x\pm \Delta{x}, y\pm \Delta{y}]$. These concepts are illustrated in Figure \ref{fig:symbols}. In our framework, there are three phases involved to achieve the scene recovery:structure estimation, structure propagation, and remaining part filling.

\subsection{Structure Estimation}
In this phase, we estimate the potential structure in $\Omega$ by finding all the possible edges. This procedure can be further decomposed into two steps: \emph{Contour Detection in $\Phi$} and \emph{Curve Generation in $\Omega$}.

\subsubsection{ Contour Detection in $\Phi$}
We first segment the region $\Phi$ by using \emph{gPb Contour Detector} \cite{Pablo_11}. It is based on the idea of computing the oriented gradient signal $G(x, y, \theta)$ on the four channels of its transformed image: brightness, color \emph{a}, color \emph{b} and texture channel. $G(x, y,  \theta)$ is the gradient signal, where $(x, y)$ indicates the center location of the circle mask that is drawn on the image and $\theta$ indicates the orientation.  The \emph{gPb Detector} is composed of two important components: \emph{mPb Edge Detector} and \emph{sPb Spectral Detector} \cite{Pablo_11}. We apply linear combination on \emph{mPb} and \emph{sPb} (factored by $\beta$ and $\gamma$) according to the gradient ascent on F-measure:

\begin{equation}
gPb(x, y, \theta) = \beta\cdot mPb(x, y, \theta) + \gamma \cdot sPb(x, y, \theta)
\end{equation}

Thus a set of edges in $\Phi$ can be retrieved via \emph{gPb}. However, these edges are not in close form and have classification ambiguities. To solve this problem, we use the \emph{Oriented Watershed Transform}\cite{Pablo_11} and \emph{Ultrametric Contour Map}\cite{Arbel_06} (\emph{OWT-UCM}) algorithm to find the potential contours by segmenting the image into different regions. The output of \emph{OWT-UCM} is a set of different contours $\{C_i\}$ and their corresponding boundary strength levels $\{\mathcal{L}_i\}$ as Figure \ref{fig:contour a} shows.

\subsubsection{Curve Generation in $\Omega$}
After obtaining the contours $\{C_i\}$ from the above procedure, salient boundaries in $\Phi$ can be found by traversing $\{C_i\}$. Our method for generating the curves in $\Omega$ is based on the assumption: for the edges on the boundary in $\Phi$ that intersects with the $\partial\Omega$, it either ends inside $\Omega$ or passes through the missing region $\Omega$ and exits at another point of $\partial\Omega$. Below is our algorithm for identifying the curve segments in $\Omega$:

%\subsection{Broken Edge Matching Algorithm and Reconstruction}
\begin{algorithm}[h]
\caption{
  %Optimal algorithm for the VASP problem (
  {\bf {Identifying curve segments in $\Omega$}}}
\label{alg1}
\small
\begin{algorithmic}
\REQUIRE {Construct curve segments in $\Omega$.}
\ENSURE {The generated curves have smooth transition between known edges.}
\end{algorithmic}
\begin{algorithmic}[1]
  \STATE Initial $\mathbf{t} = 1.0$
  \STATE For $\mathbf{t} = \mathbf{t} - \Delta\mathbf{t}$
  \STATE \qquad if $ \exists e \in \{\mathcal{C}\}: E \cap \partial\Omega \neq \emptyset$
  \STATE  \qquad \qquad Insert $e$ into $\{E\}$
  \STATE End if $\mathbf{t} < \delta_T $
  \STATE Set $\mathbf{t} = \mathbf{t}_0$, retrieve all the contours in $\{C_i\}$ with $L_i > \mathbf{t}$
  \STATE Obtain $<\phi_{x1}, \phi_{x2}>$ for each $E_x$
  \STATE $\mathbf{DP}$ on $\{<\phi_{01}, \phi_{02}>, <\phi_{11}, \phi_{12}>, ...  \}$ to find optimal pairs
  from the list.
  \STATE According to the optimal pairs, retrieve all the corresponding edge-pairs: $\{(E_{x1}, E_{x_2}), (E_{x_3}, E_{x_4}), ...)\}$.
  \STATE Compute a transition curve $\mathcal{C}_{st}$ for each $(E_{s}, E_{t})$.
\end{algorithmic}
\end{algorithm}

In algorithm \ref{alg1}, it has three main parts: (a) collect all potential edges  $\{E_x\}$ in $\Phi$ that hits $\partial\Omega$; (b) identify  optimal edge pairs $\{(E_s, E_t)\}$ from $\{E_x\}$; (c) construct a curve $\mathcal{C}_{st}$ for each edge pair $(E_s, E_t)$.

{\bf Edges Collection:}  The output of \emph{OWT-UCM} are contours sets $\{\mathcal{C}_i\}$ and their corresponding boundary strength levels $\{\mathcal{L}_i\}$. Given different thresholds $\mathbf{t}$, one can remove those contours $\mathcal{C}$ with weak $\mathcal{L}$. Motivated by this, we use the \emph{Region-Split} scheme to gradually demerge the whole $\Phi$ into multiple sub-regions and extract those salient curves. This process is carried out on lines 1-9: at the beginning the whole region $\Phi$ is considered as one contour; then iteratively decrease $\mathbf{t}$ to let potential sub-contours $\{\mathcal{C}_i\}$ faint out according the boundary strength; Every time when any edges $e$ from the newly emerged contours $\{\mathcal{C}\}$ were detected of intersecting with $\partial\Omega$, they are put into the set $\{E\}$.

{\bf Optimal Edge Pairs:} the reason of identifying edge pairs is based on the assumption if an edge is broken up by $\Omega$, there must exist a pair of corresponding contour edges in $\Phi$ that intersect with $\partial\Omega$. To find the potential pairs  $\{(E_s, E_t)\}$ from the edge list $\{E_x\}$, we measure the corresponding enclosed regions similarities. The neighboring regions $<\phi_{x1}, \phi_{x2}>$ which is partitioned by the edge $E_s$  are used to compare with the corresponding regions of another edge $E_t$. This procedure is described on lines $7-9$ of the algorithm \ref{alg1}. Each neighboring region is obtained by lowing down the threshold value $\mathbf{t}$ to faint out more detailed contours as Figure \ref{fig:contour b} shows.

%\begin{figure}
%    \centering
%    \subfigure[]
%    {
%        \includegraphics[width=6.4cm, height = 5.2cm]{pic/demo_5.eps}
%        \label{fig:contour a}
%    }
%    \subfigure[]
%    {
%        \includegraphics[width=6.4cm, height = 5.2cm]{pic/demo_7.eps}
%        \label{fig:contour b}
%    }
%    \caption{Curve Reconstruction in Missing Area $\Omega$  (a) Find the contours from original image (b) Extract edges from the contours}
%    \label{fig:contour}
%\end{figure}

 To compute the similarity between regions, we use the \emph{Jensen-Shannon divergence} \cite{Lamberti_08} method that works on the color histograms:

\begin{equation}
d(H_1, H_2) = \sum_{i = 1}^n \{H_1^i\cdot log\frac{2\cdot H_1^i}{H_1^i + H_2^i} + H_2^i\cdot log\frac{2\cdot H_2^i}{H_2^i + H_1^i} \}
\end{equation}

where $H_1$ and $H_2$ are the histograms of the two regions $\phi_{1}, \phi_{2}$; $i$ indicates the index of histogram bin. For any two edge $(E_{s}, E_{t})$, the similarity between them can be expressed as:

\begin{equation}
M(E_s, E_t) = \frac{||L_s - L_t||}{L_{max}}\cdot\min\{d(H_{si}, H_{ti}) + d(H_{sj}, H_{tj})\}
\end{equation}

$i$ and $j$ are the exclusive numbers in $\{1, 2\}$, where 1 and 2 represent the indices of the two neighboring regions in $\phi$ around a particular edge. The $L_{max}$ is the max value of the two comparing edges' strength levels. The first multiplier is a penalty term for big difference between the  strength levels of the two edges. To find the optimal pairs among the edge list, dynamic programming is used to minimize the global distance: $\sum_{s, t}M(E_s, E_t)$, where $s \neq t$ and $s, t \in \{0, 1, ..., size(\{E_i\})\}$. To enhance the accuracy, a maximum constraint is used to limit the regions' difference: $d(H_1, H_2) < \delta_H$. If the individual distance is bigger than the pre-specified threshold $\delta_H$, the corresponding region matching is not considered. In this way, it ensures if there are no similar  edges existed, no matching pairs would be identified.

%There is a special case in the edge matching: straight lines. We notice that in many structured images, straight lines are often intersect with each other, e.g. perpendicular lines. Though based on the above matching algorithm, the edges of these lines do not meet the pairing conditions. They could be still parts of salient structures if the corresponding $L$ is bigger than a certain threshold. Our strategy for these edges is to continue the straight line in $\Omega$ and it will end when it crosses another straight line or one of some other matching edges. The ways to determines straight lines are based on the equation [6].

%
%\begin{figure}[htb]
%\centerline{\epsfig{figure=pic/demo_9.eps,height = 5.2cm, width=6.4cm}}
%\caption{Edges Matching and Curve Reconstruction}\label{fig:angle}
%\end{figure}

{\bf Generate Curves for each $(E_s, E_t)$ :} we adopt the idea of fitting the clothoid segments with polyline stoke data first before generating a curve \cite{McCrae_09}. Initially, a series of discrete points along the two edges $E_s$ and $E_t$ are selected, denoted as $\{p_{s0}, p_{s1}, ..., p_{sn}, p_{t0}, p_{t1}, ..., p_{tm}\}$. These points have a distance with each other by a pre-specified value $\Delta_d$. For any three adjacent points $\{p_{i-1}, p_{i}, p_{i+1}\}$, the corresponding curvature $k_i$ could be computed according to \cite{Mullinex_07}:

\begin{equation}
k_i = \frac{2\cdot det(p_i - p_{i - 1}, p_{i + 1} - p_i)}{||p_i - p_{i-1}||\cdot||p_{i + 1} - p_i||\cdot||p_{i+1} - p_{i -1}||}
\end{equation}

Combining the above curvature factors, a sequence of polyline are used to fit these points. The polylines are expected to have a possibly small number of line segments while preserving the minimal distance against the original data.  Dynamic programming is used to find the most satisfied polyline sequence by giving a penalty for each additional line segment. A set of clothoid segments can be derived corresponding to each line segment. After a series rotations and translations over the clothoid, a final curve $\mathcal{C}$  is obtained by connecting each adjacent pair with $\mathbf{G}^2$ continuity \cite{McCrae_09}. Figure \ref{fig:angle} demonstrates the curve generation result.

\subsection{ Structure Propagation:}
After the potential curves are generated in $\Omega$, a set of texture patches, denoted as $\{\Psi_0, \Psi_1, ... \}$, need to be found from the remaining region $\Phi$ and placed along the estimated curves by overlapping with each other with a certain proportion. Similar to \cite{sun_05}, an energy minimization based method is proposed in a \emph{Belief Propagation} (BP) framework. However, we have different definitions for the energy and message passing functions. The details are in the algorithm \ref{alg3}.

%\subsection{Adaptive Patch Algorithm}
\begin{algorithm}[h]
\caption{
  %Optimal algorithm for the VASP problem (
  {\bf {\emph{BP} Propagation}}  Algorithm   }
\label{alg3}
\small
\begin{algorithmic}
\REQUIRE {Render the texture for each patch $\Psi_i$ in $\Omega$ along the estimated structures.}
\ENSURE {Find the best matching patches while ensuring the global coherence and consistency.}
\end{algorithmic}
\begin{algorithmic}[1]
  \STATE For each curve $\mathcal{C}$ in $\Omega$, define a series of \emph{anchor points} on it, $\{\mathbf{a}_i, |i = 1\rightarrow n\}$.
  \STATE Collect exemplar-texture patches $\{\hat{\Psi}_{t_i}\}$ in $\Phi$, where $t_i\in[1, m]$
  \STATE Setup a factor graph $\mathcal{G} = \{\mathcal{V}, \mathcal{E}\}$ based on $\{\mathcal{C}\}$ and $\{\mathbf{a}_i\}$
  \STATE Defining the energy function $\mathbf{E}$ for each $\mathbf{a}_i$: $\mathbf{E}_i(t_i)$, where $t_i$ is the index in $[1, M]$.
  \STATE Defining the message function $\mathbf{M}_{ij}$ for each edge $\mathcal{E}$ in $\mathcal{G}$, with initial value $\mathbf{M}_{ij} \leftarrow 0$
  \STATE Iteratively update all the messages $\mathbf{M}_{ij}$ passed between $\{\mathbf{a}_i\}$
  \STATE \qquad $\mathbf{M}_{ij} \leftarrow \min_{a_i}\{ \mathbf{E}_i(t_i) + \mathbf{E}_{ij}(t_i, t_j) + \sum_{k\in \mathcal{N}(i), k\neq j}\mathbf{M}_{ki}\}$
  \STATE end until $\Delta \mathbf{M}_{ij} < \delta$, $\forall i, j$ (by Convergence)
  \STATE Assign the best matching texture patch from $\{\hat{\Psi_t}\}$ for each $\mathbf{a}_i$ that $\arg \min_{[T, R] }\{\sum_{i\in \mathcal{V}}\mathbf{E}_i(t_i) + \sum_{(i, j)\in \mathcal{E}}\mathbf{E}_{ij}(t_i, t_j)\}$. Here $T$ is an $n$ dimensional vector $[t_1, t_2, ..., t_n]$, where $i \in [1, n]$; $R$ is also an n dimensional vector $[r_1, r_2, ..., r_n]$ with each element representing the orientation of source patch $\hat{\Psi}_{t_i}$.
\end{algorithmic}
\end{algorithm}

In the algorithm, the \emph{anchor points} are evenly distributed along the curves with an equal distance from each other $\Delta d$. These points represent the center where the patches $\{\Psi_i\}$ ($l\times l$) are synthesized, as shown in Figure \ref{fig:symbols}. In practice, we define $\Delta{d} = \frac{1}{4}\cdot l$. The $\{\hat{\Psi}_t\}$ is the source texture patches in $\Phi$. They are chosen on from the neighborhood around $\mathbf{\partial\Omega}$. For the factor graph building, we consider each $\mathbf{a}_i$ as a vertex $\mathcal{V}_i$ and $\mathcal{E}_{ij} = \mathbf{a}_i\mathbf{a}_j$, where $i$, $j$ are the two adjacent points.

In previous works \cite{sun_05}\cite{Criminisi_04}, each $\Psi_i$ have the same orientation as $\hat{\Psi}_{t_i}$, which limits the varieties in the source texture.  Noticing that different patch orientations could produce different results, we introduce a scheme called \emph{Adaptive Patch} by defining a new formulas for $\mathbf{E}$ and $\mathbf{M}$ in the structure propagation.

Traditionally, the node energy $\mathbf{E}_i(t_i)$ is defined as the \emph{Sum of Square Difference}(SSD) by comparing the known pixels in each patch $\Psi_i$ with the candidate corresponding portion in $\hat{\Psi}_{t_i}$. But this method limits the salient structure directions. Instead of using SSD on the two patches, a series of rotations are performed on the candidate patch before computing the similarity. Mathematically, $\mathbf{E}_i(t_i)$ can be formulated as:

\begin{equation}
\mathbf{E}_i(t_i) = \alpha_\lambda\cdot P\cdot\sum||\Psi_i - \dot{R}(\theta)\cdot\hat{\Psi}_{t_i}||_{\lambda}^2
\end{equation}

Where $\dot{R}$ represents different rotations on the patch $\hat{\Psi}_{t_i}$. Since the size of a patch is usually small, the rotation can be specified with an arbitrary number of angles. In our experiment, it is specified as $\theta \in \{0, \pm\frac{\pi}{4}, \pm\frac{\pi}{2}, \pi\}$. The parameter $\lambda$ represents the number of known pixels in $\Psi_i$ that overlap with the rotated patch $\hat{\Psi}_{t_i}$. $P$ is a penalty term, the more number of overlapping pixels, the higher of similarity is assigned. So we use $P$ to discourage the patches with smaller number of sharing pixels. Here, the percentage is expressed as $P = \frac{\lambda}{l^2}$ ($l$ is the length of $\Psi$). $\alpha_\lambda$ is the corresponding normalized scalar. Thus the best matching patch $\hat{\Psi}$ is represented by two factors: index $t_i$ and rotation $R_i$.

In a similar way, the energy $\mathbf{E}_{ij}(t_i, t_j)$ on each edge $\mathcal{E}_{ij}$ can be expressed as:

\begin{equation}
\mathbf{E}_{ij}(t_i, t_j) = \alpha_{\lambda}\cdot P\cdot\sum||\Psi_i(t_i, \theta_{t_i}) - \Psi_j(t_j, \theta_{t_j})||_\lambda^2
\end{equation}

Here $i$ and $j$ are the indices of the two adjacent patches in $\Omega$. A penalty scheme is applied to the similarity comparison. The two parameters for $\Psi_i$ indicate the index and rotation for the source patches in $\{\hat{\Psi}_{t_i}\}$. The messages propagation is derived from the results of the above energy functions. We adopt a similar method as \cite{sun_05}, where the message $M_{ij}$ passes by patches $\Psi_i$ is defined as:

\begin{equation}
M_{ij} = \mathbf{E}_i(t_i) + \mathbf{E}_{ij}(t_i, t_j)
\end{equation}

Through iterative updating on the \emph{BP} graph, an optimal decision of $\{t_i\}$ for the patches in $\{\Psi_i\}$ is made by minimizing the nodes' energy. This principle can be formulated in the definition below:

\begin{equation}
\hat{t}_i = \arg \min_{t_i}\{\mathbf{E}_i(t_i) + \sum_k M_{ki}\}
\end{equation}

Where $k$ is one of the neighbors of the patch $\Psi_i$: $k \in\mathcal{N}(i)$.  $\hat{t}_i$ is the optimal index for the matching patch. To achieve minimum global energy cost, dynamic programming is used. Each assignment for  $\Psi_i$ or $\mathbf{a}_i$ is considered as a stage. In each stage, the choices of $\hat{\Psi}_{t_i}$ represent different states. The edge $\mathcal{E}_{ij}$ represents the transit cost from state $\hat{\Psi}_{t_i}$ at stage $i$ to state $\hat{\Psi}_{t_j}$ at stage $j$. Starting from $i = 0$, an optimal solution is achieved by minimizing the total energy $\xi_i(t_i)$ from last step:

\begin{equation}
\xi_i(t_i) = \mathbf{E}_i(t_i) + min\{\mathbf{E}_{ij}(t_i, t_j) + \xi_{i-1}(t_{i-1})\}
\end{equation}

 where $\xi_i(t_i)$ represents a set of different total energy values at current stage $i$. In the situation of multiple intersections among curves $\mathcal{C}$, we adopted the idea in \cite{sun_05}, where readers can refer for further details.

\subsection{ Remaining Part Filling:}
After the curves are generated in $\Omega$, we fill the remaining regions by using the exemplar-based approach in \cite{Criminisi_04}. The $\partial\Omega$ is getting smaller and smaller by spreading out the known pixels $\Phi$ in a certain order. To enhance the accuracy, all the pixels in the above generate patches along the estimated curves are assigned with a pre-computed confidence value based on the  confidence updating rule in \cite{Criminisi_04}.

%% file: exp.tex
\section{Experiments}
\label{sec:exp}
%\vspace{-0.1in}

\begin{figure}[!htbp]
    \centering
    \subfigure[]
    {
        \includegraphics[width=3.0cm, height = 3.0cm]{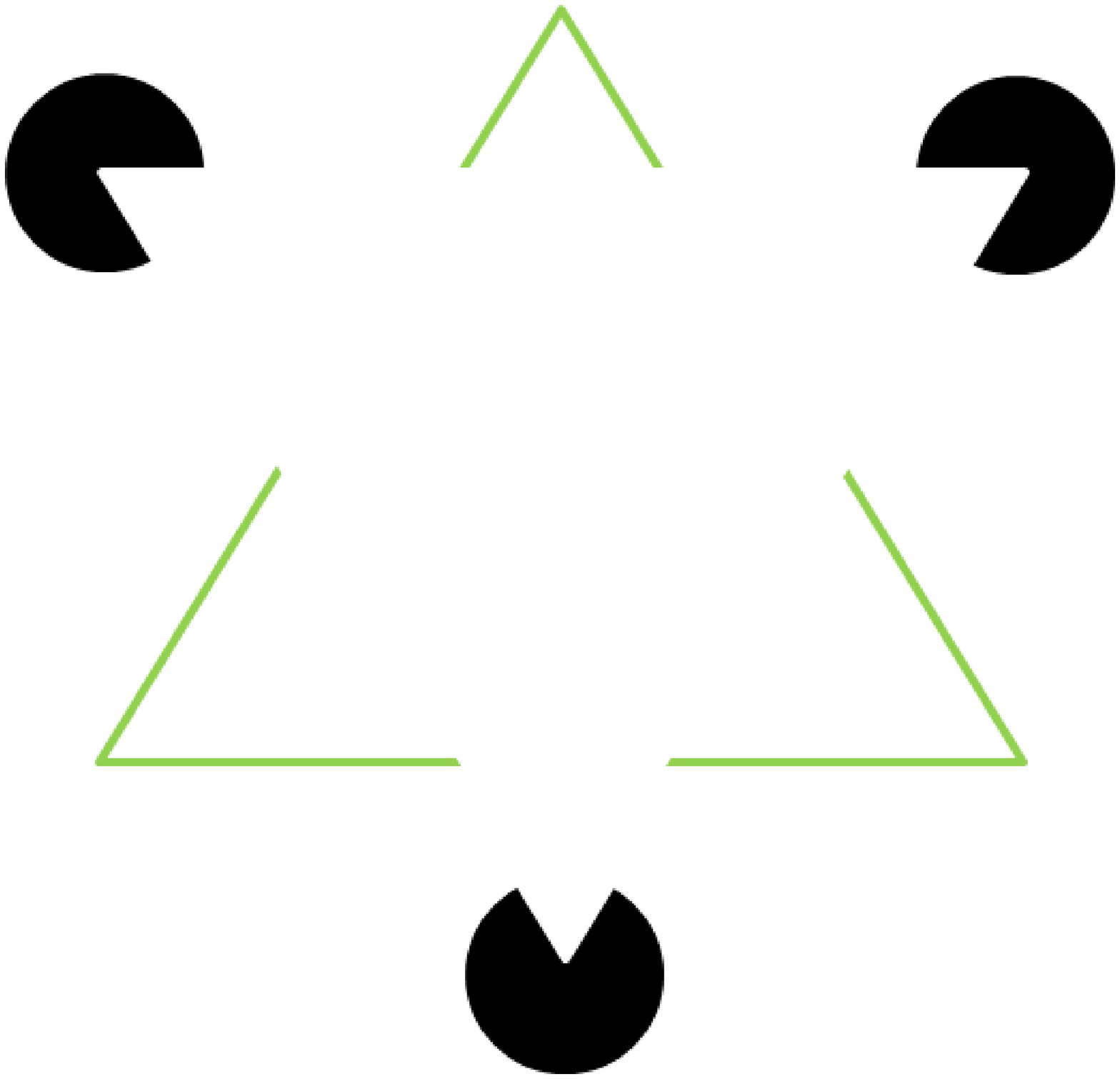}
        \label{fig:triangle a}
    }
    \subfigure[]
    {
        \includegraphics[width=3.0cm, height = 3.0cm]{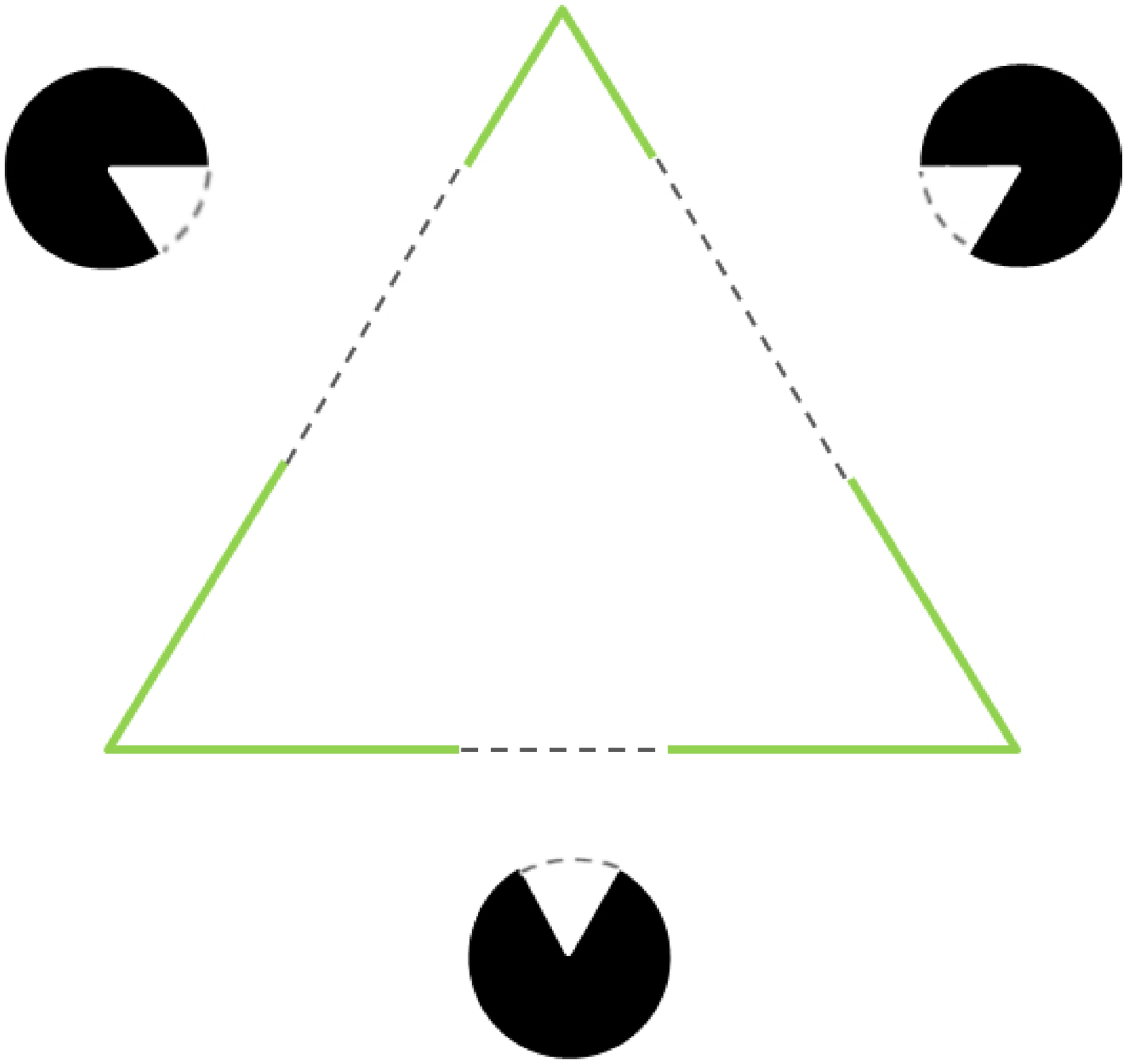}
        \label{fig:triangle b}
    }
    \\
    \subfigure[]
    {
        \includegraphics[width=3.0cm, height = 3.0cm]{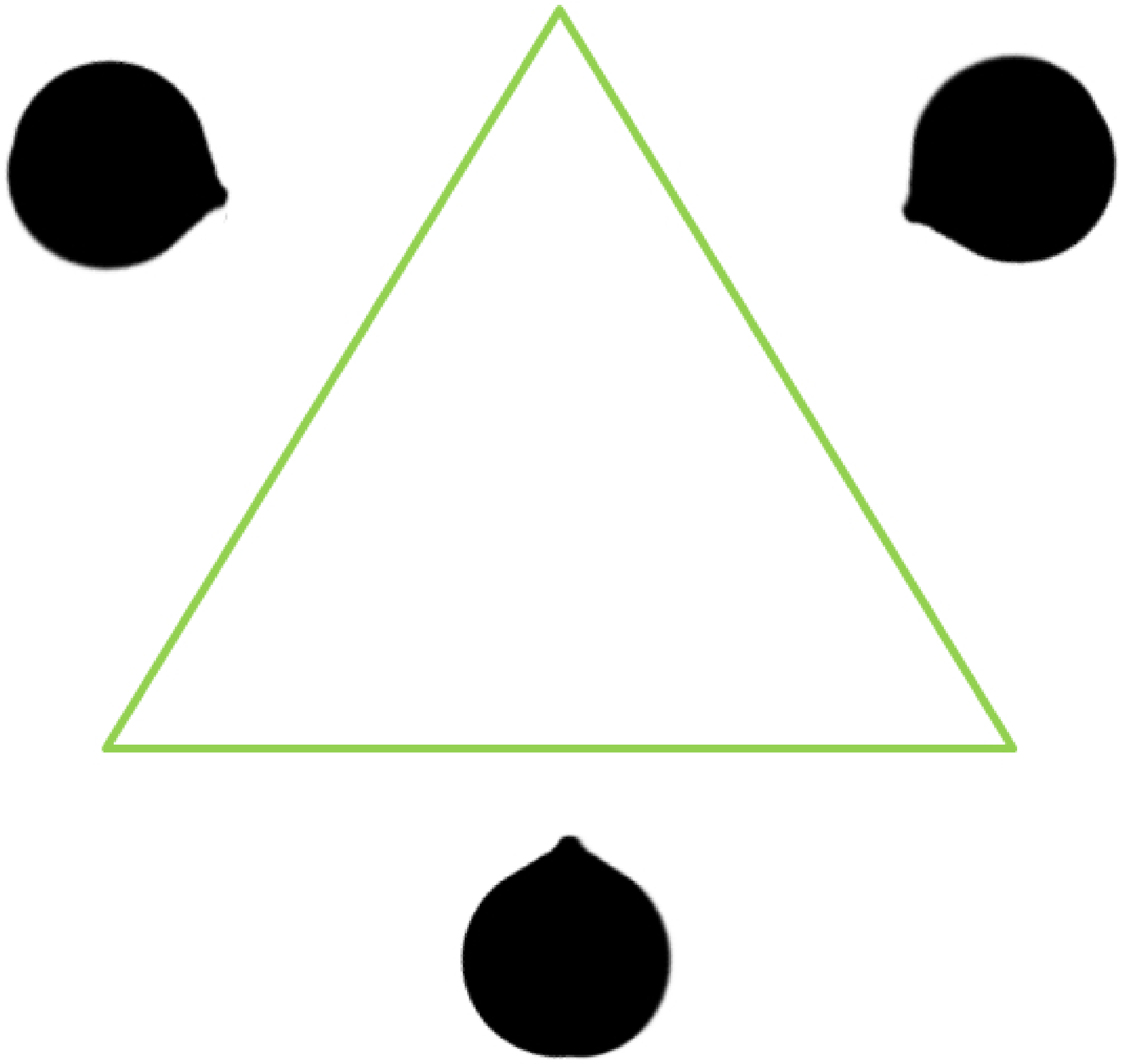}
        \label{fig:triangle c}
    }
    \subfigure[]
    {
        \includegraphics[width=3.0cm, height = 3.0cm]{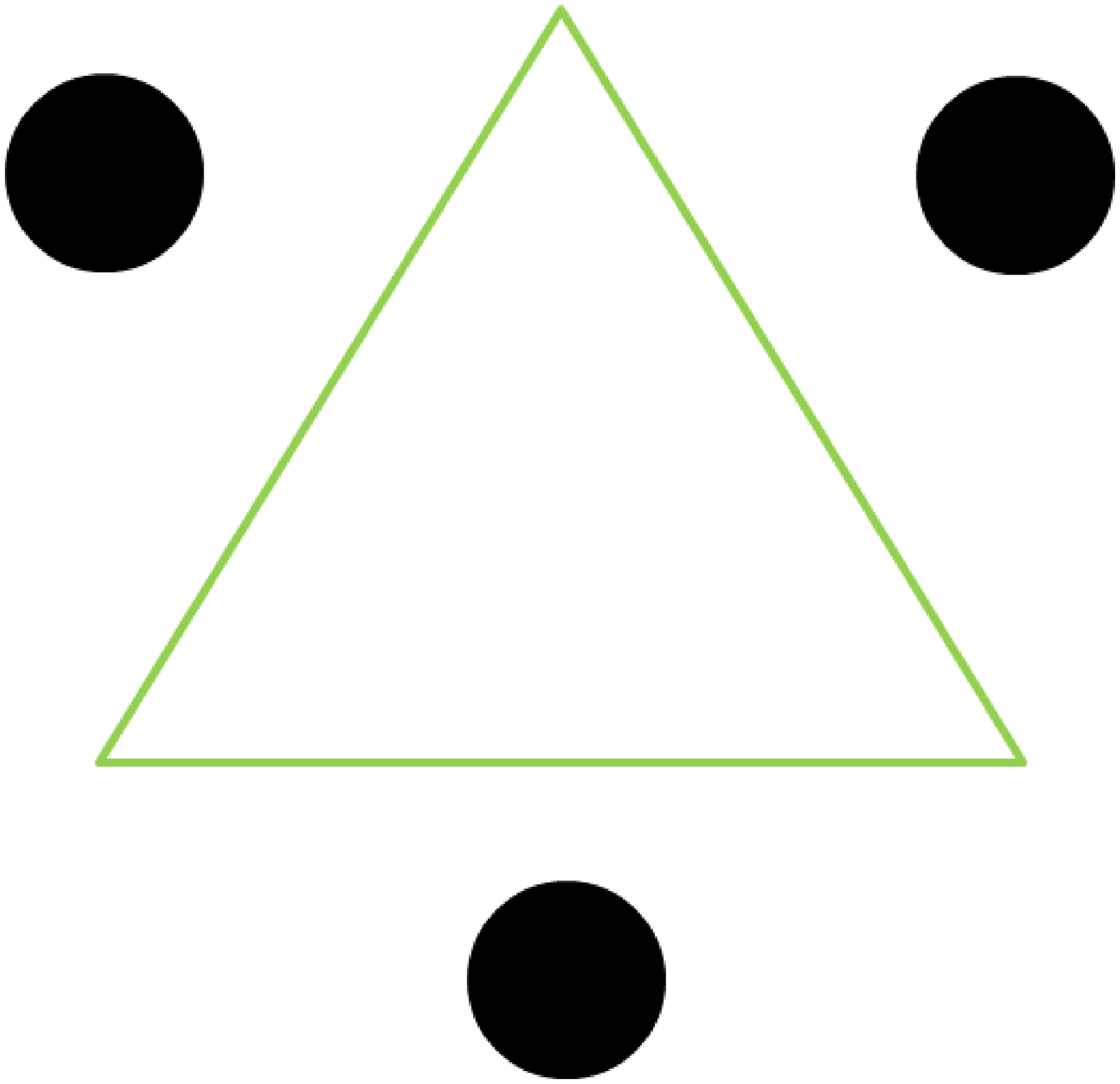}
        \label{fig:triangle d}
    }
    \caption{Kanizsa Triangle Experiment  (a) The original Image (b)Curve reconstruction for the missing region $\Omega$ (c) Result by Criminisi's method (d) Our result.}
    \label{fig:triangles}
\end{figure}

%To verify the accuracy of our algorithm, both structure and histogram based analysis are employed to quantify the errors. Our approach is as follows:  several images are collected and mask each by the region with salient structures. These regions are considered as $\Omega$, which need to be filled by image in-painting. Then a set of new images are generated with the masked region filled. These images are used to compare with the original images by computing the histograms and edge differences in the masked regions $\Omega$. The
%
%As the quality of image in-painting rely much on the subjects' attitudes. Below we demonstrated several images generated by our algorithms, which user can refer to evaluate and compare with others' work.
In our experiments, we first evaluate our proposed approach in terms of structure coherence by comparing our result with the one  in \cite{Criminisi_04} that works on the well-known \emph{Kanizsa triangle}. As shown in Figure \ref{fig:triangle a}, the white triangle in the front is considered as the occluded region $\Omega$ that needs to be removed. First, a structure propagation is carried out based on the detected edges along $\partial\Omega$. The dash lines in Figure \ref{fig:triangle b} indicate the estimated potential structures in $\Omega$. Texture propagation is applied to the rest of the image based on the confidence and isophote terms. One can notice both the triangle and the circles are well completed in our result Figure \ref{fig:triangle d} comparing with Criminisi's method in Figure \ref{fig:triangle c}.

\begin{figure*}[!htbp]
    \centering
    \subfigure[]
    {
        \includegraphics[width=4.25cm, height = 3.7cm]{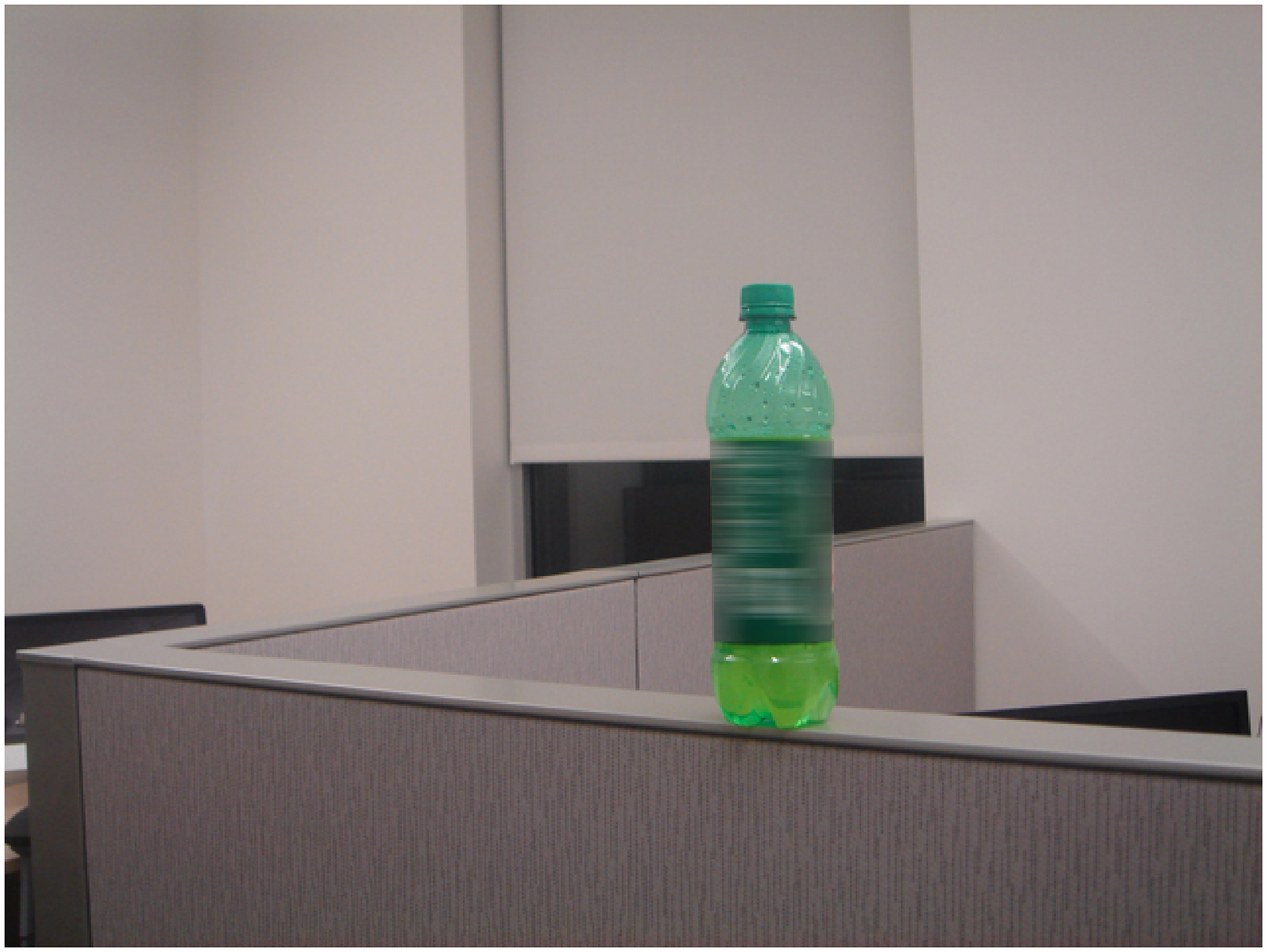}\hspace{-0.3cm}
        \label{fig:faces a}
    }
    \subfigure[]
    {
        \includegraphics[width=4.25cm, height=3.7cm]{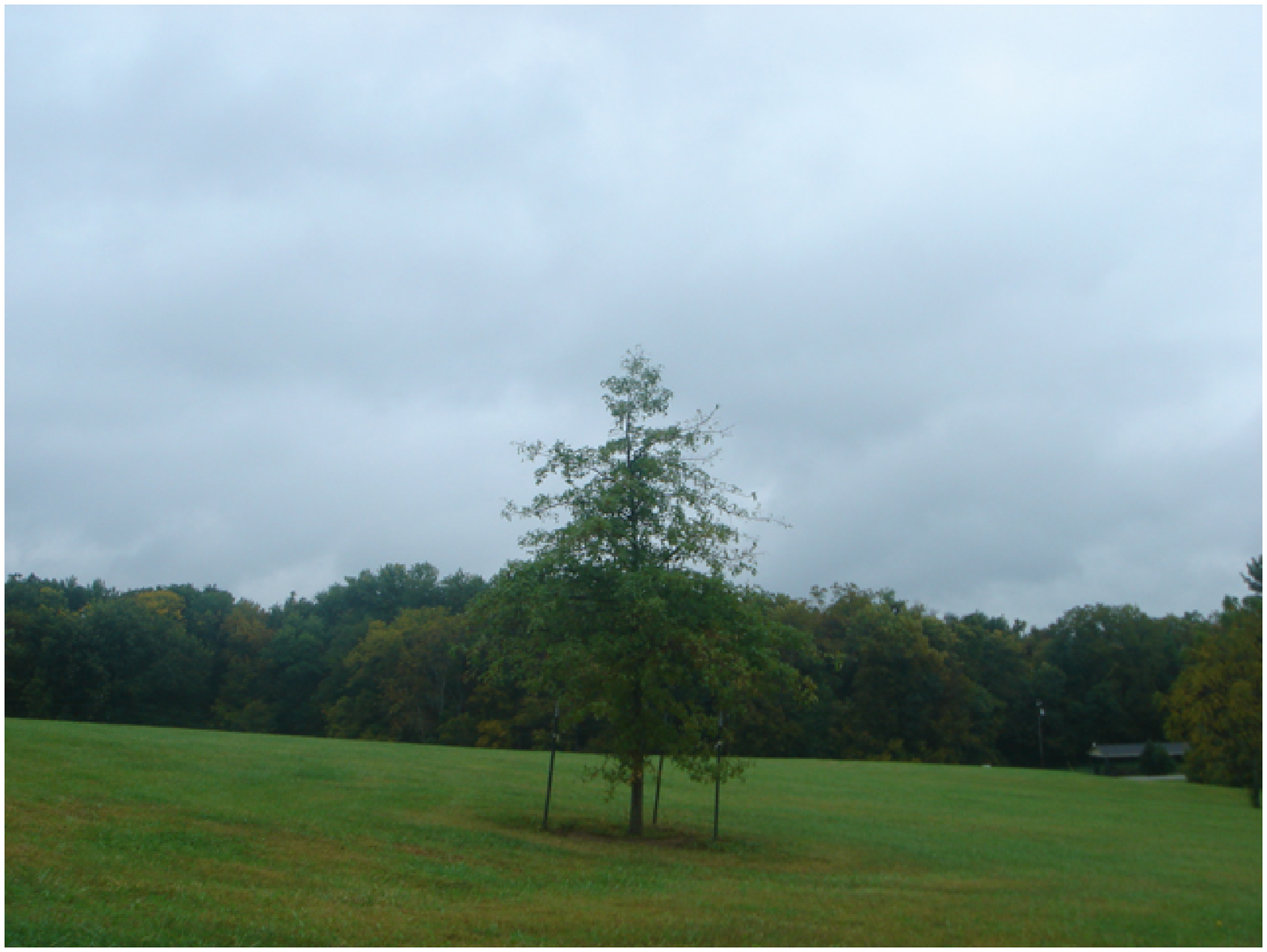}\hspace{-0.3cm}
        \label{fig:tree1}
    }
     \subfigure[]
    {
        \includegraphics[width=4.25cm, height=3.7cm]{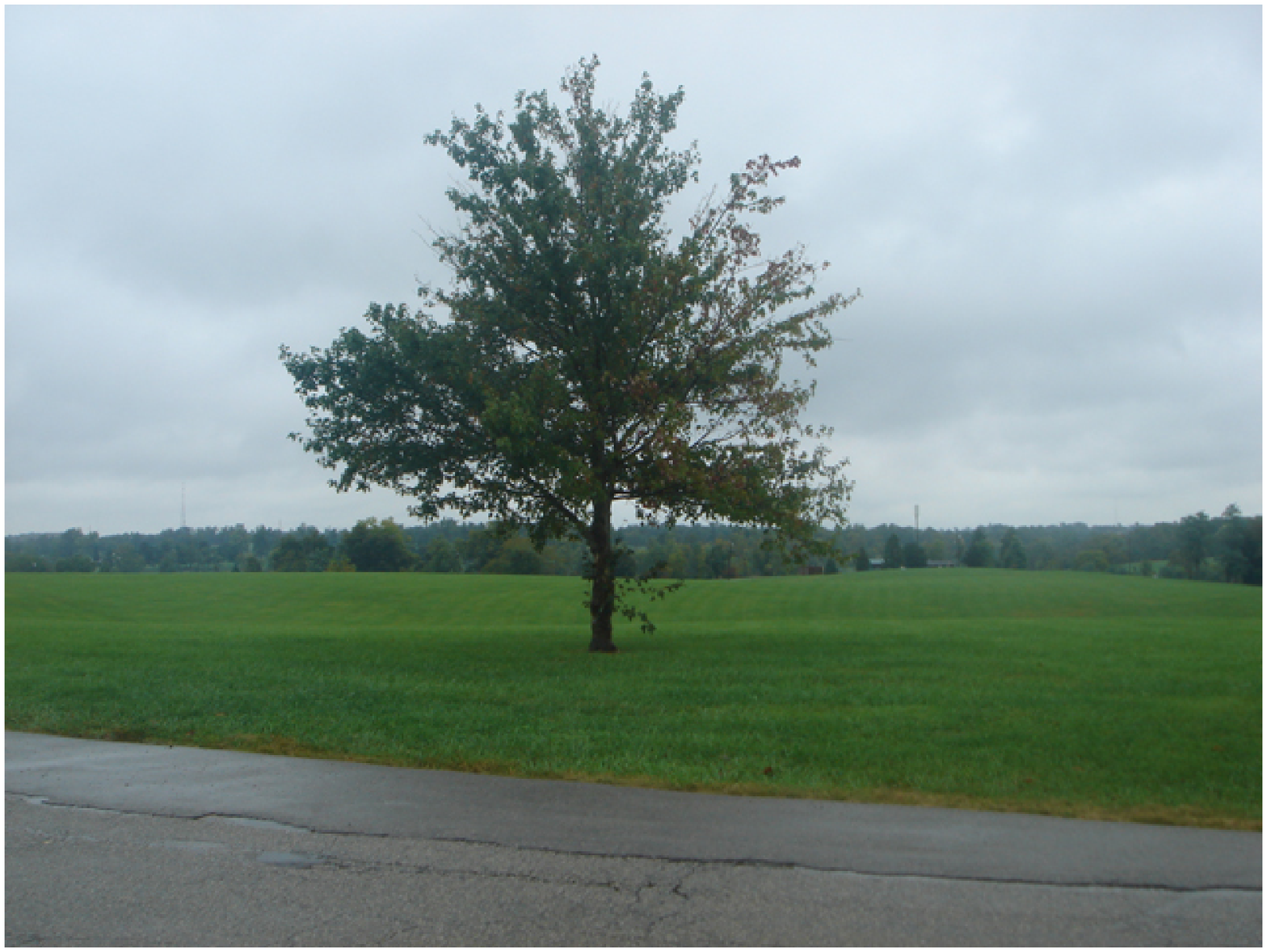}\hspace{-0.3cm}
        \label{fig:tree2}
    }
    \subfigure[]
    {
        \includegraphics[width=4.25cm, height=3.7cm]{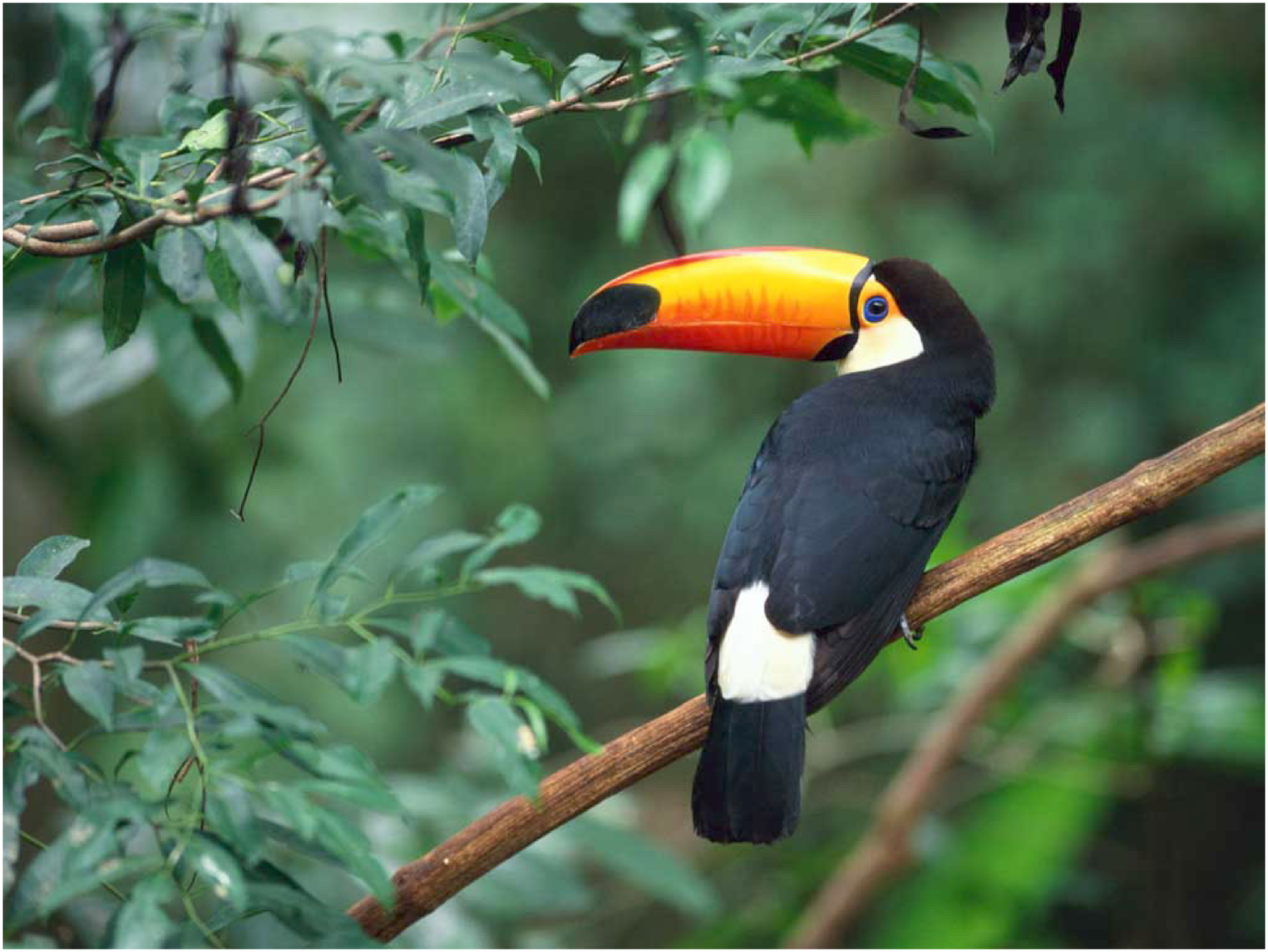}
        \label{fig:bird}
    }
    \subfigure[]
    {
        \includegraphics[width=4.25cm, height = 3.7cm]{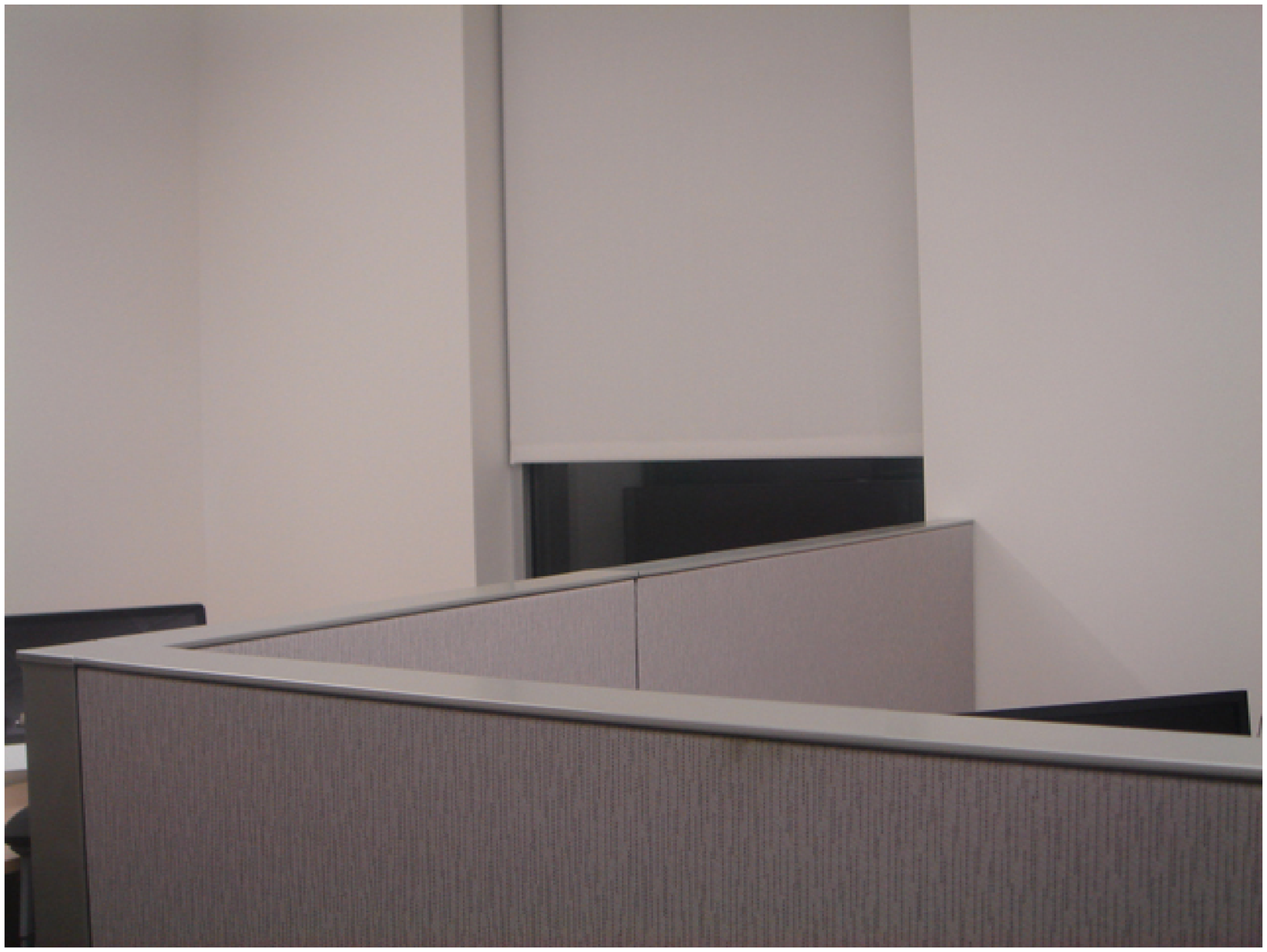}\hspace{-0.3cm}
        \label{fig:faces b}
    }
    \subfigure[]
    {
        \includegraphics[width=4.25cm, height=3.7cm]{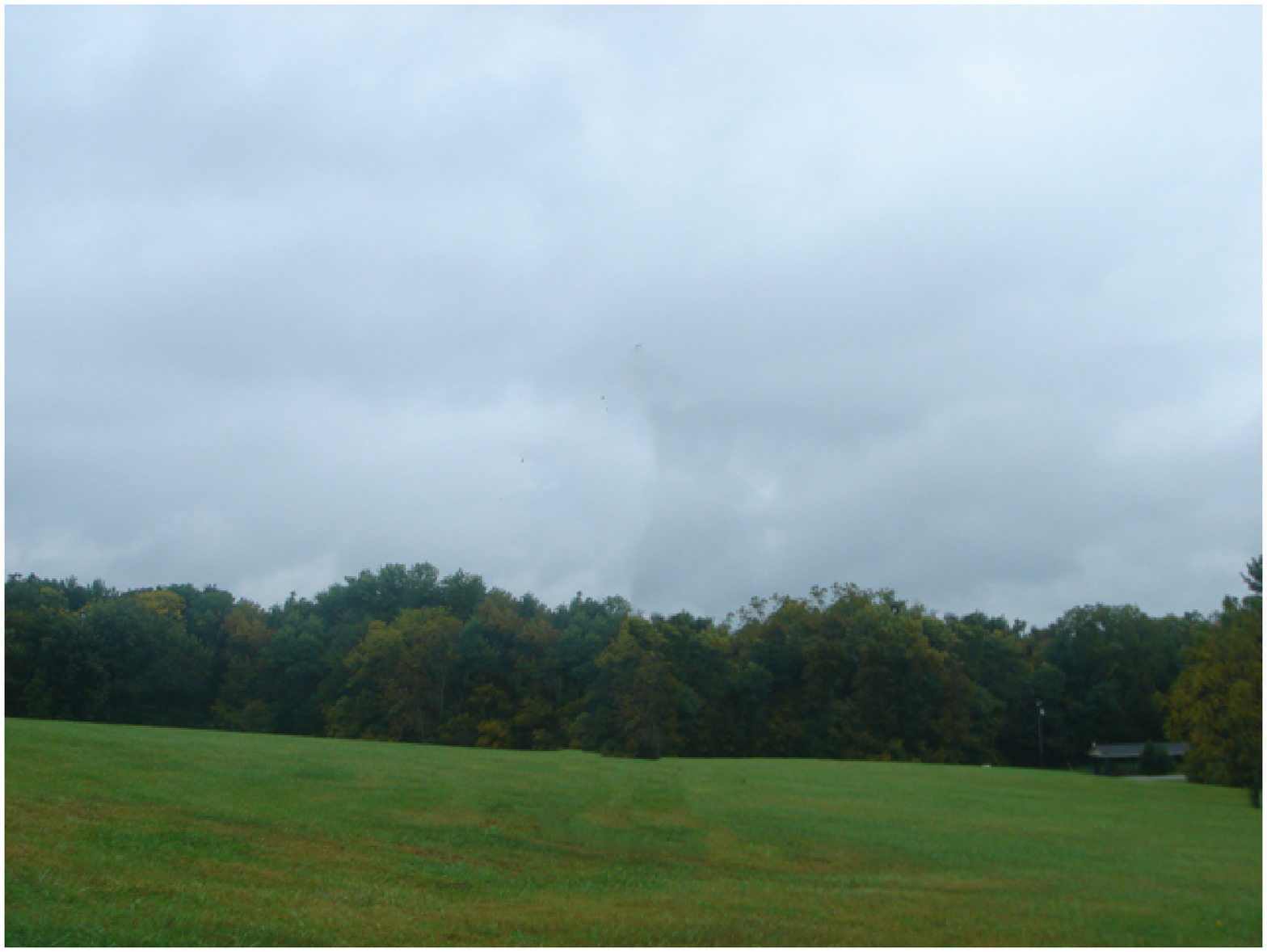}\hspace{-0.3cm}
        \label{fig:tree_1}
    }
    \subfigure[]
    {
        \includegraphics[width=4.25cm, height=3.7cm]{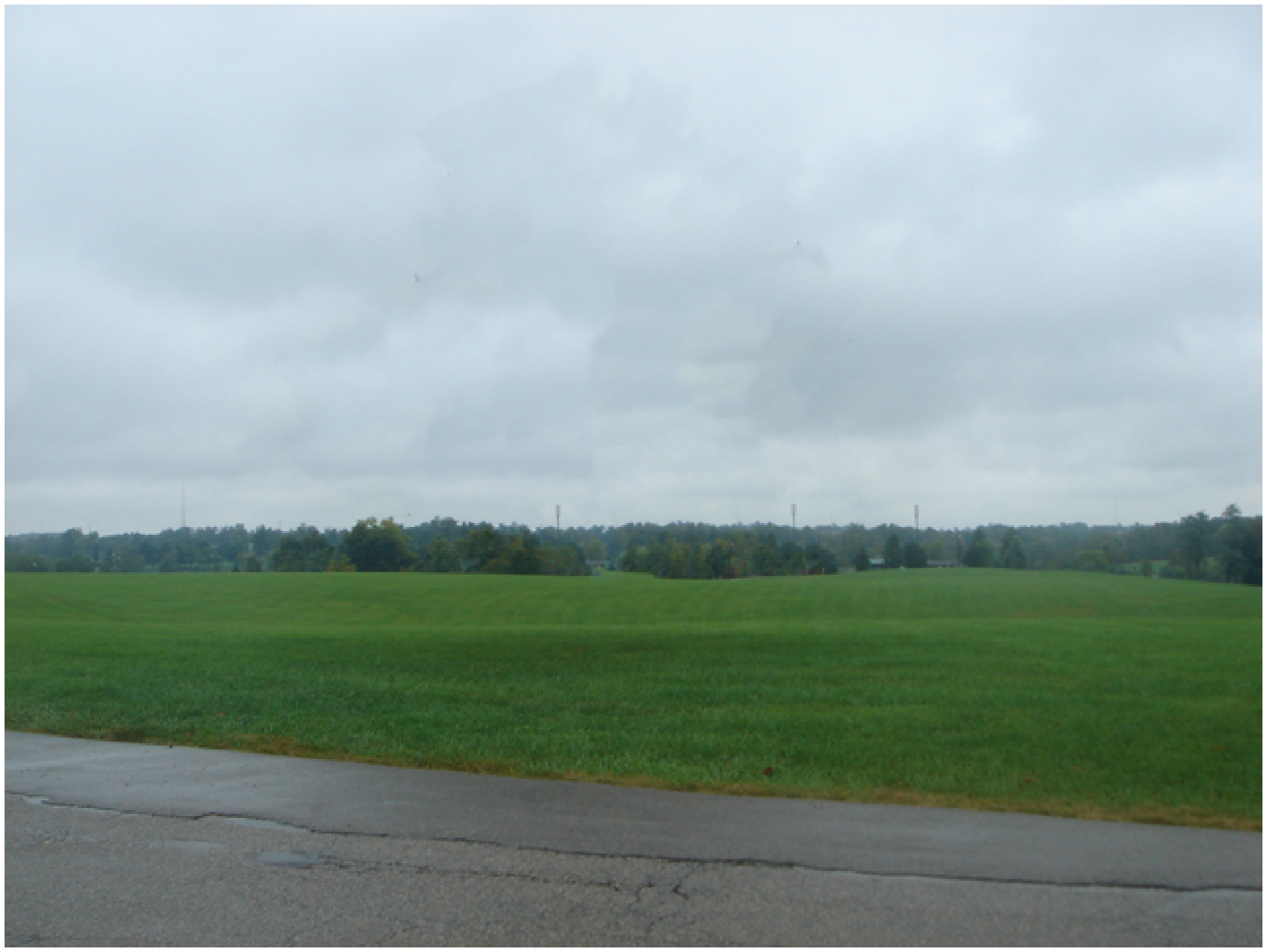}\hspace{-0.3cm}
        \label{fig:tree_2}
    }
    \subfigure[]
    {
        \includegraphics[width=4.25cm, height=3.7cm]{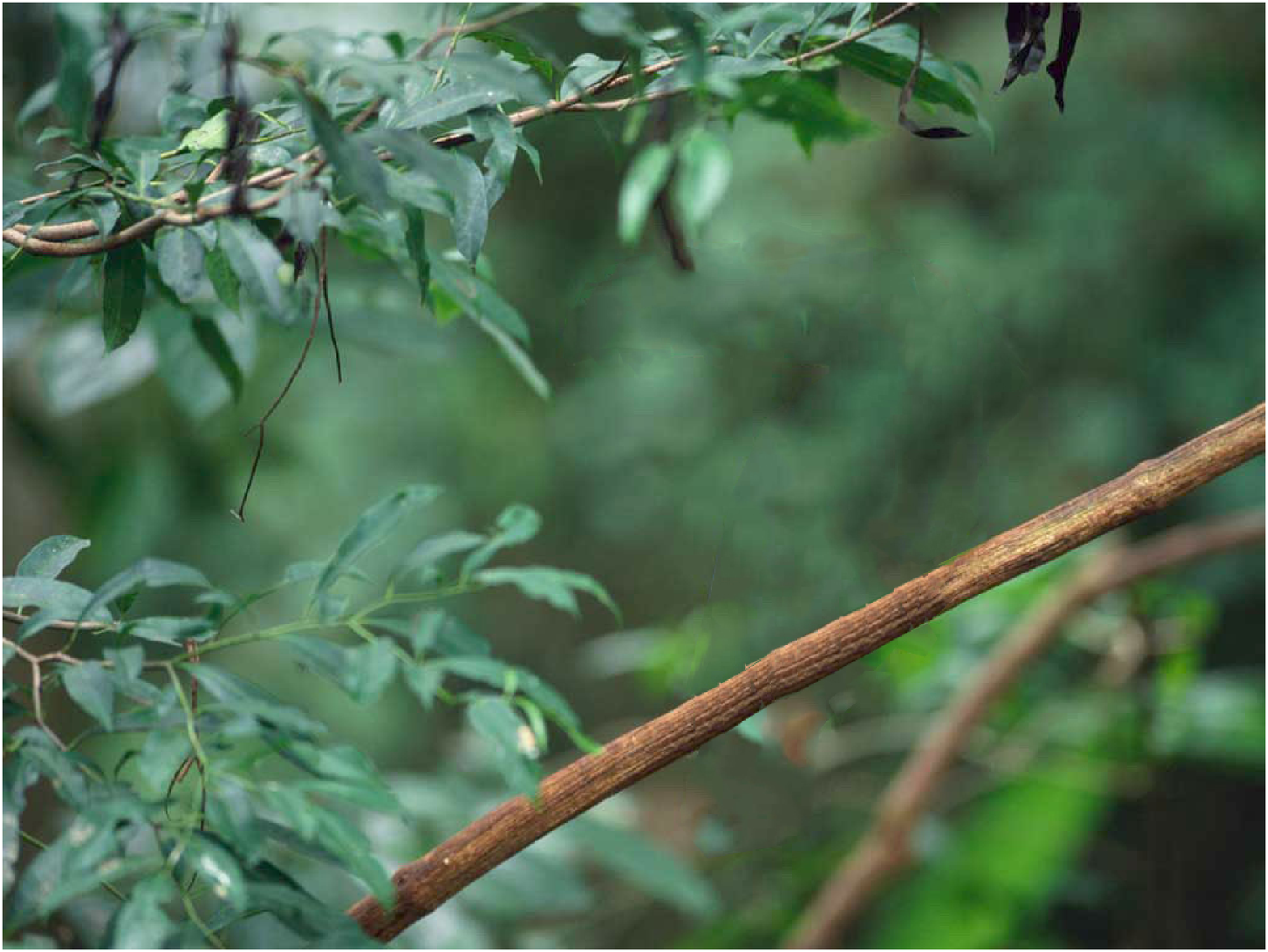}
        \label{fig:bird_2}
    }

    \caption{Scene reconstruction results on different settings: the first row shows the original images; the second row shows the corresponding result images.}
    \label{fig:tree1}
\end{figure*}

To further demonstrate the performance, a set of images are used for scene recovery: ranging from indoor environment to natural scenes. Figure \ref{fig:faces b} shows an indoor case where highly structured patterns often present, such as the furniture, windows, walls. The green bottle on the office partition is successfully removed while preserving the remaining structure. In this example, three pairs of edges are identified and connected by the corresponding curves that are generated in the occluded region $\Omega$. Figure \ref{fig:tree_2} and \ref{fig:tree_1} show the results of removing trees in the nature scenes. Several curves are inferred by matching the broken edges along $\partial\Omega$ and maximizing the continuity. We can notice the three layers of the scene (sky, background trees, and grass land) are well completed. In Figure \ref{fig:bird_2}, it shows a case that a perching bird is removed from the tree. Our structure estimation successfully completes the tree branch with smooth geometric and texture transitions. 

%% file: q_con.tex
\section{Conclusion}
\label{sec:con}

In this paper, we present a novel approach for foreground objects removal while ensuring structure coherence and texture consistency. The core of our approach is using structure as a guidance to complete the remaining scene.
%, which demonstrates promising results. 
This work would benefit a wide range of applications especially for the online massive collections of imagery, such as photo localization and scene reconstructions. 
%By removing foreground objects, the matching accuracy can be dramatically improved as the corresponding features  are only extracted from the static scene rather than those moving objects. The generated views are more realistic because the foreground pixels are not involved in any image transformation and geometric estimation. Furthermore, 
Moreover, this work is applied to privacy protection by removing people from the scene. 
%As our future work, we will apply this object removal technique to scene reconstruction applications that experiment directly on the internet photo streams.

%\cite{Che_95} \cite{Ashikhmin_01} \cite{Bonet_97} \cite{Efros_01} \cite{Efros_99} \cite{Freeman_00} \cite{Heeger_95} \cite{Hertzmann_01} \cite{Igehy_97} \cite{c_01} \cite{Wey_00} \cite{Criminisi_04} \cite{P_01} \cite{R_02} \cite{A_02} \cite{L_01} \cite{M_00} \cite{Bertalmio_03} \cite{Drori_03} \cite{Jia_03} \cite{Masnou_98} \cite{Masnou_02} \cite{Wexler_07} \cite{Komodakis_06}